# Emulating Human Cognitive Processes for Expert-Level Medical Question-Answering with Large Language Models

Khushboo Verma[1], Marina Moore[1], Stephanie Wottrich[1], Karla Robles López[1], Nishant Aggarwal[2], Zeel Bhatt[2], Aagamjit Singh[2], Bradford Unroe[3], Salah Basheer[4], Nitish Sachdeva[5], Prinka Arora[6], Harmanjeet Kaur[6], Tanupreet Kaur[6], Tevon Hood[7], Anahi Marquez[7], Tushar Varshney[8], Nanfu Deng[9], Azaan Ramani[9], Pawanraj Ishwara[10], Maimoona Saeed[11], Tatiana López Velarde Peña[12], Bryan Barksdale[13], Sushovan Guha[14,15], Satwant Kumar[16]*

[1]Department of Neurology, Dell Medical School at the University of Texas, Austin, Texas, USA

[2]William Beaumont University Hospital, Royal Oak, Michigan, USA

[3]Austin State Hospital, Austin, Texas, USA

[4]Institute of Mental Health, University of Nottingham, Nottingham, United Kingdom

[5]Department of Medicine, Sub Divisional Hospital, Dera Bassi, SAS Nagar, Punjab, India

[6]Department of Psychiatry, Government Medical College and Hospital, Chandigarh, India

[7]Department of Pediatrics, Dell Medical School at the University of Texas, Austin, Texas, USA

[8]Department of Psychiatry, Bezirkskrankenhaus Augsburg, University of Augsburg, Bayern, Germany

[9]Department of Internal Medicine, McGovern Medical School and The University of Texas Health Science Center, Houston, Texas, USA

[10]Division of Neurology, Sheikh Shakhbout Medical City, Abu Dhabi, United Arab Emirates

[11]Department of Pediatrics, Dr Akbar Niazi Teaching Hospital, Islamabad, Pakistan

[12]Servicio de Geriatría, Centro Médico ABC, Ciudad de México, Mexico

[13]Department of Psychiatry, Dell Medical School at the University of Texas, Austin, Texas, USA

[14]Department of Surgery, McGovern Medical School and The University of Texas Health Science Center, Houston, Texas, USA

[15]Houston Regional Gastroenterology Institute, Houston, Texas, USA

[16]NeuroReef Labs, Austin, Texas, USA

*Corresponding author: Satwant Kumar, SatwantKumar@NeuroReef.com

**Abstract**

In response to the pressing need for advanced clinical problem-solving tools in healthcare, we introduce BooksMed, a novel framework based on a Large Language Model (LLM). BooksMed uniquely emulates human cognitive processes to deliver evidence-based and reliable responses, utilizing the GRADE (Grading of Recommendations, Assessment, Development, and Evaluations) framework to effectively quantify evidence strength. For clinical decision-making to be appropriately assessed, an evaluation metric that is clinically aligned and validated is required. As a solution, we present ExpertMedQA, a multispecialty clinical benchmark comprised of open-ended, expert-level clinical questions, and validated by a diverse group of medical professionals. By demanding an in-depth understanding and critical appraisal of up-to-date clinical literature, ExpertMedQA rigorously evaluates LLM performance. BooksMed outperforms existing state-of-the-art models Med-PaLM 2, Almanac, and ChatGPT in a variety of medical scenarios. Therefore, a framework that mimics human cognitive stages could be a useful tool for providing reliable and evidence-based responses to clinical inquiries.

Several use cases for large language models (LLMs) have been reported in the healthcare setting in recent months[1–4]. In the medical field, ChatGPT and Med-PaLM 2 have been noted for their promising applications[5,6]. Among the notable results of ChatGPT is its passing grades in the United States Medical Licensing Examinations, with GPT-4 showing substantial improvements over GPT-3.5[7,8]. Med-PaLM 2, a specialized variant of PaLM 2, has achieved state-of-the-art results, close to the expertise of human clinicians in addressing health-related user queries[6]. When ChatGPT's output was compared with the responses of physicians, the model's output was favored for its quality and empathetic communication, as determined by medical professionals[4]. While current LLMs hold promising potential, there is a discernible disparity between their anticipated and actual performances. Although current LLMs exhibit quantitative progress, they fall short of making qualitative leaps, often providing general rather than patient-tailored information[1,5,9,10]. Furthermore, they have difficulty applying intricate medical knowledge in a comprehensive way[11–13]. Moreover, current benchmarks do not adequately reflect the complexities inherent in clinical decision-making, demonstrating the need for robust and clinically aligned evaluation metrics[1,4,14]. The absence of uncertainty indicators in the outputs of the models also creates significant challenges[10]. Without these indicators, evaluating the reliability and accuracy of the provided information becomes a daunting task, which could lead to the dissemination of misleading information and biases[7,15]. In the healthcare sector, where accuracy and reliability are critical, this shortfall is particularly alarming. The need for enhanced LLMs characterized by a comprehensive understanding of medical knowledge and its precise application is imperative in order to bridge these gaps. In addition, these enhanced models should include uncertainty indicators to minimize misinformation risks, as well as comprehensive evaluation benchmarks to truly reflect the nuances of clinical decision-making[1]. As a result of these vital improvements, these improved LLMs can become indispensable tools in the healthcare sector, providing valuable support to clinicians and ultimately contributing to improved patient care.

We developed BooksMed to address these identified challenges in the application of LLMs within the healthcare sector. BooksMed is a novel framework designed for advanced clinical problem-solving. Unlike existing models, BooksMed excels in generating evidence-based responses, uniquely quantifying the strength of evidence using the GRADE (Grading of Recommendations, Assessment, Development, and Evaluations) framework, thereby providing reliable assistance in medical scenarios[16,17].

BooksMed framework is designed to simulate human-like complex problem-solving [18–20]. In this framework, a series of models is used, each finely tuned to a specific stage of the decision-making process, similar to how clinicians comprehend, analyze, synthesize, and apply medical knowledge in their daily practice[17,21,22]. Within BooksMed, each model plays a pivotal role: some are proficient at interpreting clinical scenarios, while others are proficient at evaluating the strength and reliability of medical evidence. As a result of this structured, collaborative approach, BooksMed is able to provide responses that are not only accurate and comprehensive but also reliable. BooksMed consistently outperformed its competitors, including Google's Med-PaLM series[6,23], ChatGPT[8], and Almanac[14] in a variety of medical scenarios.

Next, we introduce ExpertMedQA, a more robust and reflective benchmark for LLMs in healthcare. Featuring open-ended, expert-level clinical questions, this dataset can rigorously test the capabilities of advanced LLMs, such as BooksMed. ExpertMedQA requires LLMs to not only have a foundational understanding of clinical literature but also engage in deep and critical evaluations, mirroring the intricate and nuanced decision-making processes encountered in real-life healthcare settings[17,21,22]. ExpertMedQA questions reflect the depth and complexity of daily decisions made by healthcare professionals, offering a realistic and challenging benchmark for LLMs. Based on the validation by a diverse group of medical professionals worldwide, ExpertMedQA establishes a new, rigorous standard for evaluating LLMs, serving as an invaluable tool for ongoing assessment and improvement of models such as BooksMed.

In terms of medical question-answering and clinical decision support, this article provides two key contributions (1) Introduction of BooksMed: BooksMed offers evidence-based responses that are able to assist healthcare professionals in making well-informed clinical decisions. (2) Development and validation of ExpertMedQA: ExpertMedQA is the first validated benchmark that examines the expert-level performance of language models. Therefore, it could be a catalyst for future AI-driven healthcare research and future model development. The article provides solutions to existing challenges, establishing the basis for future exploration and development in AI-assisted healthcare.

# Methods

**ExpertMedQA**
Using open-ended medical questions for analysis, we explore the potential of novel large language model-based frameworks in clinical medicine. This task requires a comprehensive understanding of the latest developments in biomedical literature, as well as an ability to examine the strength and validity of scientific research evidence.

*Development of ExpertMedQA*
We have developed and validated ExpertMedQA, a novel benchmark characterized by open-ended, expert-level clinical questions, which bridge this gap by requiring not only an understanding of the most recent clinical literature but also an analysis of the strength of the evidence presented. From current treatment guidelines to open-ended discussions requiring knowledge and analysis based on current clinical research studies, this dataset covers a wide range of topics.

We validated ExpertMedQA using ten distinct axes to emphasize the robustness and reliability of the dataset. The axes are as follows:

1. **Expertise Required to Answer**: Ranging from 1 (accessible to the General Public) to 5 (demands expertise at the level of a Field Expert).
2. **Clarity of the Question**: Spanning from 1 (Very Confusing/Not Clear) to 5 (Very Clear/Straightforward).
3. **Depth of Knowledge Required**: Classified from 1 (requires Surface Level Knowledge) to 5 (necessitates In-depth Understanding).
4. **Relevance to Current Trends/Research**: Assessed from 1 (Outdated/Not Relevant) to 5 (Highly Pertinent to Current Practices/Research).
5. **Specificity of the Question**: Ranging from 1 (Very General/Can be applied to many topics) to 5 (Highly Specific to a Particular Topic/Area).
6. **Potential for Critical Thinking**: Spanning from 1 (Purely Factual/Recall-based) to 5 (mandates Critical Analysis/Deep Thought).
7. **The Breadth of the Topic Covered**: Classified from 1 (Focuses on a Very Narrow Area) to 5 (Encompasses a Broad Spectrum of the Field).
8. **Originality of the Question**: Evaluated from 1 (Commonly Asked/Typical Question) to 5 (offers a Unique/Original Perspective).
9. **Importance in Clinical or Research Setting**: Ranging from 1 (Rarely Relevant) to 5 (Frequently Encountered/Key Concept).
10. **Applicability for Assessment**: Spanning from 1 (Not Suitable for Testing Knowledge) to 5 (Ideal for Gauging Expertise/Understanding).

An overview of the dataset is depicted in **Table 1**. **Extended Data** provides the complete ExpertMedQA dataset along with responses generated by BooksMed, highlighting the dataset's diversity and complexity, and providing a comprehensive overview of dataset questions.

**Table 1: Overview of ExpertMedQA**

| Medical Specialty | Number of Questions |
|---|---|
| Pediatrics | 20 |
| Internal Medicine | 20 |
| Psychiatry | 20 |
| Neurology | 20 |
| **Total** | **80** |

**Framework for Clinical Evaluation**

*Consortium of Evaluators*
The evaluation of BooksMed responses within the ExpertMedQA dataset was conducted by a global consortium of evaluators. This group was diverse, consisting of clinical practitioners, residents, and fellows specializing in Neurology, Internal Medicine (IM), Psychiatry, and Pediatrics. The evaluators hailed from various countries, including the United States of America, India, United Kingdom, Germany, Mexico, Pakistan, and the United Arab Emirates, providing a wide range of perspectives and expertise. Each field had representation from four to seven evaluators, ensuring a comprehensive and varied assessment of BooksMed responses.

*Evaluation Criteria*
The evaluators were tasked with following explicit instructions for the assessment process, utilizing a Likert scale ranging from 1 (Not Very) to 5 (Very Much) for the following criteria:

1. **Factual Accuracy of Answer**: Evaluators scrutinized the factual correctness embedded within the answers, ensuring alignment with recognized medical knowledge, and validating the supporting citations for accuracy and relevance.
2. **Adequacy of the Answer**: The depth, breadth, and relevance of each answer were examined to ascertain whether it adequately addressed the posed question in its entirety.
3. **Correctness in Formatting**: The presentation and layout of answers were reviewed, taking into account the logical structuring, coherence of content, use of markdown formatting for enhanced clarity, and adherence to citation formatting standards.
4. **Clarity & Precision**: Each answer's clarity in articulating complex medical information was assessed, with a focus on the precision of terminologies and phrasing, ensuring they were apt for an expert reader audience.
5. **Citation Relevance and Appropriateness**: The relevance and appropriateness of citations within the answers were evaluated, with a spotlight on the credibility of the sources and the efficacy with which they supported the attributed statements.

*Advantages of the Evaluation Approach*
The evaluative approach adopted offered two significant advantages:

1. **Bias Mitigation**: The diverse backgrounds of the evaluators, spanning different geographical regions, disease epidemiology and medical training protocols, allowed for an unbiased and nuanced assessment of BooksMed responses[24,25].
2. **Consensus Building**: Engaging a diverse set of evaluators facilitated the building of a consensus that mirrors the global medical community's viewpoints, resulting in a robust and universally applicable evaluation of BooksMed responses[24,25].

The same pool of evaluators conducted comparative evaluations of different models.

**BooksMed Framework**

Drawing inspiration from human problem-solving paradigms, expert decision-making methodologies, and intelligence research, BooksMed emulates human-like complex problem-solving (CPS) processes[18,26,27]. Unlike conventional systems relying on a single LLM, BooksMed deploys distinct fine-tuned models at each problem-solving phase **(Table 2)**. **Figure 1** illustrates the various stages of the problem-solving process as well as the various components of the BooksMed system. All models under BooksMed have a shared core architecture, ensuring consistency and synergy in their functions. For reproducibility and objective comparison of results, BooksMed uses OpenAI's 'GPT-3.5-turbo' model.

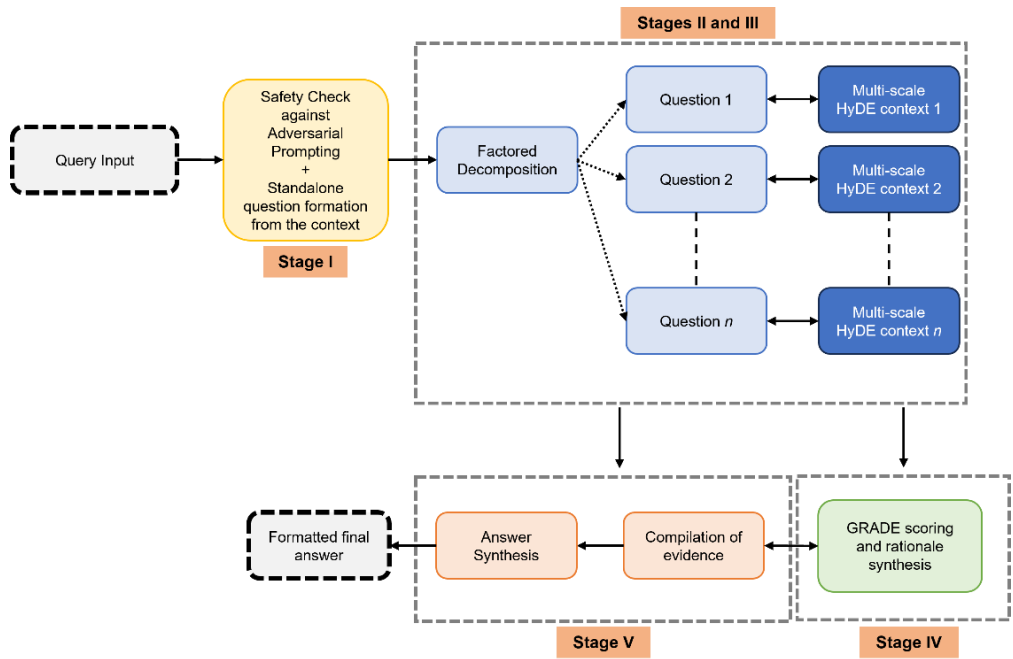

**Figure 1: Overview of BooksMed Framework**. Illustrated are the five stages of the BooksMed framework. **Stage I:** Problem Identification and Representation involves ambiguity resolution and precise problem framing. **Stage II:** Knowledge Acquisition phase ensures access to extensive information for informed solution formulation. **Stage III:** Strategy Formulation and Planning generates and organizes solution strategies. **Stage IV:** Monitoring and Evaluation tracks

and assesses proposed solutions against benchmarks. **Stage V:** Reflection, Iteration, and Continuous Improvement involves cycles of learning and strategy refinement. Each stage utilizes fine-tuned models working synergistically, with a shared core architecture. **Table 2** outlines each stage in detail.

**Table 2: Overview of the BooksMed complex problem-solving stages and components.**

|   | Description | Corresponding BooksMed Framework Components |
|---|---|---|
| **Stage I**: Problem Identification and Representation | Initial ambiguity, if any is clarified through precise problem identification and semantic representation, forming a foundation for solution strategies. | 1. *Safety Measures*: At this stage, safety checks against adversarial prompting are crucial, safeguarding the model from potential adversarial attacks[28].<br>2. *Formulation of a standalone question:* Understanding and formulating a descriptive standalone question using any additional context that is provided. |
| **Stage II**: Knowledge Acquisition | Ensuring access to a wealth of information through continuous learning or external knowledge integration. This phase is pivotal for informed solution formulation. | *A multiscale context retrieval and ranking method*. BooksMed's context databases are optimized for rapid indexing and vector storage. The system encodes content semantically with dense vectors, enabling efficient retrieval using a cosine distance metric. To enhance document retrieval efficiency, BooksMed implements the following strategies:<br>1. **Multiscale Context Retrieval**: By deploying vector databases with varying token lengths, BooksMed ensures a more flexible and nuanced approach to document retrieval.<br>2. **Employment of Multiple Embedding Models**: To mitigate bias in similarity searches, BooksMed utilizes a variety of embedding models, including:<br>   1. 'text-embedding-ada-002' by OpenAI, with an output dimension of 1,536.<br>   2. 'bge-large-en-v1' and 'bge-small-en-v1' by BAAI, with |
| | | dimensions of 1,024 and 384, respectively[29].<br>3. **Fine-Tuning of Embeddings**: The BGE embeddings are fine-tuned to align with the specific use case of BooksMed[29]. OpenAI embeddings are customized through a process involving multiplication with a matrix derived from supervised training on a dataset pertinent to BooksMed's use case. This process yields 'custom embeddings', emphasizing aspects of text crucial for BooksMed's functionality.<br>4. **Implementation of HyDE Method**: To fortify document retrieval capabilities, BooksMed incorporates the HyDE method[30]. This method involves generating hypothetical document embeddings (HyDE) that capture relevance patterns (albeit with potential inaccuracies) and then encoding these into embedding vectors. These vectors subsequently identify neighborhoods in the corpus of embedding space, facilitating the retrieval of similar real documents. |
| **Stage III**: Strategy Formulation and Planning | Multiple solution strategies are generated and organized coherently, addressing the problem's inherent complexity. | *Factored Decomposition*: In cognitive neuroscience literature, the concept of decomposing a complex task into subtasks has been demonstrated to enhance problem-solving abilities and enhance overall performance[18]. Similarly, in LLMs, question decomposition improves the accuracy of model-generated reasoning[31]. We implemented the factored decomposition approach. Sub-questions and answers are generated by factored decomposition, and each sub-question is answered within a new context[31]. |
| **Stage IV**: Monitoring and Evaluation | Continuous process tracking and solution evaluation against established benchmarks address multiple, often conflicting goals. | *Strength of Evidence Assessment:* Providing accurate and reliable scores based on the GRADE framework, is crucial for evaluating the quality of evidence and the strength of recommendations in healthcare[16]. |

| Stage V: Reflection, Iteration, and Continuous Improvement | The model undergoes cycles of learning, iteration, and strategy refinement based on feedback, aligning with various system dynamics. | *Answer Formulation & Final Formatting*: In-context[32] and chain-of-thought (CoT) reasoning[33] prompts are combined for answer construction, with the final answer undergoing format checks to ensure compliance with preferred standards. |
|---|---|---|

**Comparative Analysis**

We evaluated BooksMed against other state-of-the-art models to assess its comparative performance and reliability.

*Models for Comparison*

Several state-of-the-art models were compared to our model for a comprehensive appraisal. Specifically, we compared Med-PaLM and Med-PaLM 2, which are sophisticated models fine-tuned to appropriately encapsulate clinical knowledge[6,23]. Moreover, we compared it to the Almanac, a retrieval augmented LLM[14] and ChatGPT, a widely recognized model for various language processing tasks, including medical query responses[2,3,8,34]. Through these comparative analyses, we aimed to provide a systematic and impartial assessment of BooksMed's performance relative to other leading models in the field.

*Evaluation Strategy: BooksMed versus Almanac and ChatGPT*

Using the methodology described in Zakka *et al*., (2023), we evaluated BooksMed on the ClinicalQA dataset to provide a transparent and rigorous comparison[14]. Based on the identical rubric proposed by Zakka *et al*., (2023) we compared responses from model Almanac with BooksMed to a subset of the ClinicalQA dataset (n=25)[14]. Zakka *et al*., (2023) reported that Almanac outperformed ChatGPT significantly in factuality across all examined specialties, with improvements in completeness and safety[14]. Consequently, ChatGPT was excluded from the direct comparison, and only Almanac and BooksMed were evaluated on ClinicaQA. **Supplementary Table 1** describes the overview of the ClinicalQA dataset, and a subset of the questions used for the comparison. Similarly, BooksMed was evaluated against both Almanac and ChatGPT (Version as of March 23) employing the Adversarial Prompts (n=5) from Zakka *et al*., (2023)[14].

*Evaluation Strategy: BooksMed versus Med-PaLM and Med-PaLM 2 on Long-Form Answers*

Using Med-PaLM and Med-PaLM 2, Singhal *et al*., (2023) provided examples of long-form questions and their answers[6]. Using the identical axes as described by Singhal *et al*, (2023), a three-object ranking preference analysis was conducted based on long-form answers to three alternatives (BooksMed, Med-PaLM, and Med-PaLM 2).

*Statistical Evaluation*

We utilized distribution free nonparametric statistics, randomization or permutation tests. Parametric statistics were used to complement some analyses (e.g., ANOVA). The Binomial test of proportions was utilized for the analysis of five-point Likert scale responses. To account for multiple comparisons, which might increase the false discovery rate, we adjusted the p-values derived from our tests using the Benjamini and Hochberg method[35] and report them as q-values. Furthermore, the Kruskal-Wallis one-way ANOVA was utilized to examine the variations in responses across various clinical fields for each axis under evaluation. Friedman ANOVA was employed for the comparative ranking analysis involving BooksMed, Med-PaLM, Med-PaLM 2, Almanac, and ChatGPT.

# Results

*ExpertMedQA Dataset Validation*

The ExpertMedQA dataset underwent rigorous validation across ten defined axes (detailed in Methods) to ensure its reliability and comprehensiveness. This validation included multiple criteria, from the required expertise level to answer a question to its assessment applicability. For each axis, we computed the median rating on the 5-point Likert scale across all evaluators. The Spearman-Brown corrected split-half correlation coefficient ($r_{sh}$) was 0.91, indicating that we had a sufficient number of evaluators (n = 21) to obtain a reliable estimate of ratings. **Table 3** summarizes the ExpertMedQA dataset's validation results, highlighting its robustness and reliability for evaluating medical language models. The validation ratings were performed by the same 21 raters who rated the BooksMed responses. Binomial tests of proportions using the 5-point Likert scale for responses ranging from 4 to 5 yielded significant q-values (less than 0.01) in nine out of ten dimensions **(Table 3)**. The validation process confirmed that the dataset is not only comprehensive but also nuanced, capturing the subtle complexities and demands of medical knowledge and expertise, and is suitable for comprehensive testing of medical AI models.

**Table 3. Validation summary of ExpertMedQA on the ten axes.**

| Validation Dimensions [Likert Scale] | Median Ratings [95% CI] | q-Values (see text) |
|---|---|---|
| **Expertise Required to Answer** [1=General Public, 5=Field Expert] | 4.0 [3.73 – 4.26] | <0.0001 |
| **Clarity of the Question** [1=Very Confusing/Not Clear, 5 = Very Clear/Straightforward] | 4.0 [3.59 – 4.41] | <0.0001 |
| **Depth of Knowledge Required** [1 = Surface Level Knowledge, 5 = In-depth Understanding required] | 4.0 [3.62 – 4.39] | 0.0003 |
| **Relevance to Current Trends/Research** [1 = Outdated/Not Relevant, 5 = Highly Relevant to Current Practices/Research] | 4.5 [4.15 – 4.85] | <0.0001 |
| **Specificity of the Question** [1 = Very General/Can be applied to many topics, 5 = Highly Specific to a Particular Topic/Area] | 4.0 [3.73 – 4.27] | <0.0001 |
| **Potential for Critical Thinking** [ 1 = Purely Factual/Recall-based, 5 = | 4.0 [3.57 – 4.43] | 0.0576 |
| Requires Critical Analysis/Deep Thought] | | |
| **Breadth of the Topic Covered** [1 = Focuses on a Very Narrow Area, 5 = Covers a Broad Aspect of the Field] | 4.0 [3.50 – 4.50] | 0.0064 |
| **Originality of the Question** [1 = Commonly Asked/Typical Question, 5 = Unique/Original Perspective] | 4.0 [3.54 – 4.46] | 0.0064 |
| **Importance in Clinical or Research Setting** [1 = Rarely Relevant, 5 = Frequently Encountered/Key Concept] | 5.0 [4.59 – 5.00] | 0.0003 |
| **Applicability for Assessment** [1 = Not Suitable for Testing Knowledge, 5 = Ideal for Assessing Expertise/Understanding] | 5.0 [4.65 – 5.00] | <0.0001 |

*Performance Evaluation of BooksMed on ExpertMedQA*
BooksMed exhibited proficient performance in generating answers to free-form questions across diverse clinical specialties, namely, Internal Medicine, Neurology, Pediatrics, and Psychiatry. The responses generated by BooksMed were favorably rated (scores of 4 or 5) on the 5-point Likert scale across all axes tested for each specialty **(Figure 2)**. This indicates that the responses were factually accurate, adequately addressed the questions, were presented with correct formatting, were clear and precise, and utilized relevant citations. The significance of these ratings, underscored by q-values < 0.0001 (binomial test of proportions for rating in the range 4 to 5), attests the model's consistency and reliability in generating responses that align with the evaluative criteria of clinicians across different specialties. Furthermore, no significant differences were found between the median ratings across specialties (Kruskal–Wallis one-way ANOVA, q-values > 0.46), suggesting that the model consistently produces expert-level answers across specialties.

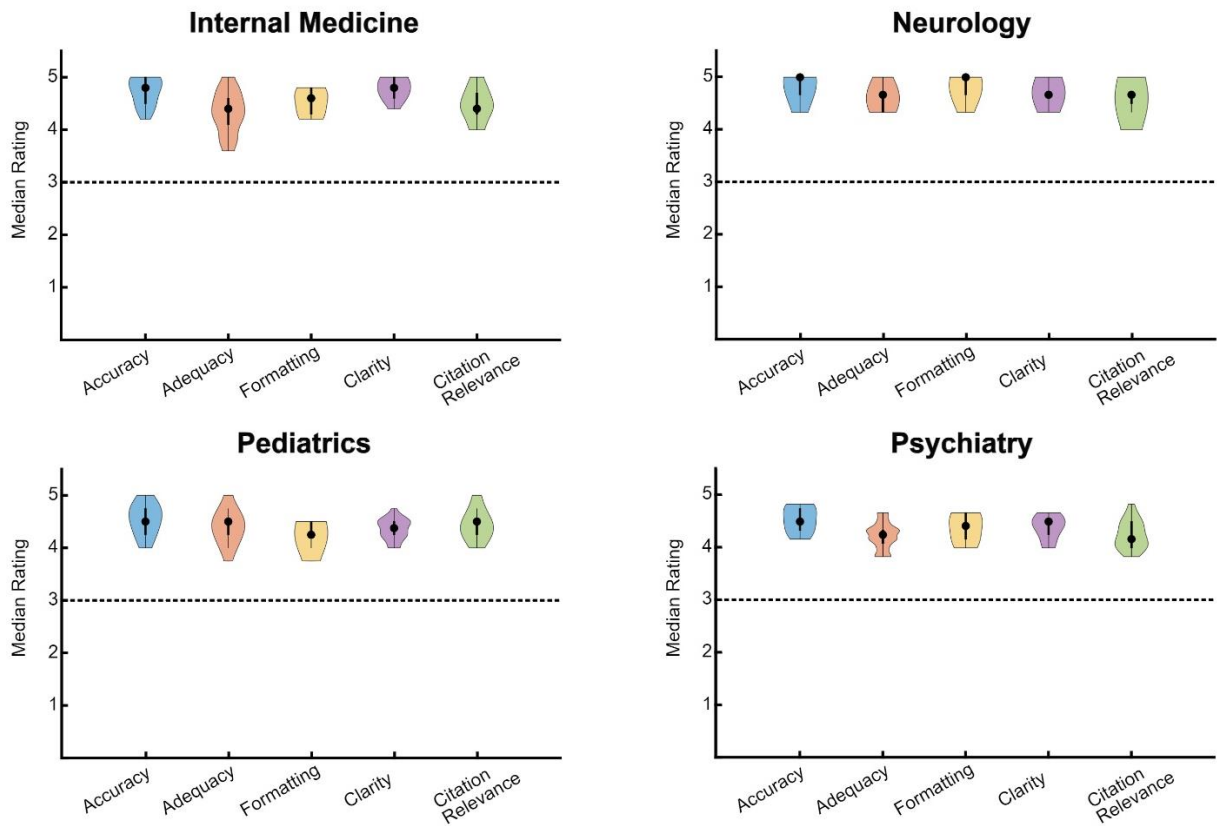

**Figure 2. Overview of BooksMed Performance on ExpertMedQA.** Distribution of median ratings across four specialties (n = 20 per specialty) for each evaluation axis: accuracy, adequacy, formatting, precision, and citation relevance. Violin plots illustrate kernel probability density, i.e., the width of the shaded area represents the proportion of data located there. An overlay boxplot indicates medians (black-filled circles) and quartiles with whiskers stretching up to the interquartile range. The stippled horizontal line indicates the neutral level on the 5-point Likert scale. All the ratings in the range of 4 to 5 on the Likert-scale across each dimension and specialty were statistically significant (q-values < 0.0001, binomial test of proportions).

**Comparison of BooksMed with Almanac and ChatGPT**

*ClinicalQA dataset*
An openly available subset of the ClinicalQA dataset (n=25, **Supplementary Table 1**) was utilized to compare BooksMed to Almanac using the methodology employed by Zakka *et al.*, (2023)[14]. On ClinicalQA dataset, BooksMed was the unanimous choice, preferred 100% of the time when considering aggregate performance **(Supplementary Figure 1)**. Examples of detailed questions and their answers for each model (BooksMed and Almanac) are available in **Supplementary Table 2**. Upon dissecting the performance on individual axes:

1. *Accuracy*: BooksMed was favored 89% of the time, indicating its reliable provision of factually accurate responses.

2. *Completeness*: With a preference rate of 96%, BooksMed showcased its ability to provide comprehensive answers that thoroughly addressed the queries posed.
3. *Safety*: BooksMed was preferred 96% of the time for its safe responses, underscoring its reliable and responsible provision of medical information.

Each axis's preference rates were statistically significant with q-values < 0.0001, as determined by the binomial test of proportions.

*Adversarial Prompts and comparison with Almanac and ChatGPT*
In response to the adversarial set of questions (n=5), BooksMed outperformed both Almanac and ChatGPT. A minimum of 87% of the evaluations across all axes showed clinicians preferring BooksMed, proving its ability to withstand adversarial and malicious prompting queries. This preference was statistically significant, with a p-value < 0.0067 obtained via Friedman ANOVA.

*Comparison of BooksMed with Med-PaLM and Med-PaLM 2*
A comparison of the preference rankings provided by three different physician raters of BooksMed, Med-PaLM, and Med-PaLM 2 is provided in **Table 4**. This provides an analytical comparison across various imperative metrics as described by Singhal *et al*, 2023 for the effectiveness of medical question-answering systems[6]. In **Supplementary Table 3**, we provide comparison questions and corresponding model-generated answers. BooksMed consistently outperforms Med-PaLM and Med-PaLM 2 in terms of aligning with scientific consensus and reading comprehension. This demonstrates BooksMed's capability to provide easy-to-understand text while remaining closely aligned with established medical knowledge and consensus. Similarly, BooksMed responses were preferred, while Med-PaLM and Med-PaLM 2 were prone to providing inaccurate information or omitting essential details. Additionally, clinicians preferred responses by BooksMed over the other two models as BooksMed responses avoided potential harm and the possibility of demographic bias. However, a clear distinction in these categories did not exist between Med-PaLM and Med-PaLM 2. The differences in model rankings are statistically significant, as evidenced by a p-value of < 0.0001 obtained through Friedman ANOVA.

**Table 4. Preference rankings of BooksMed, Med-PaLM, and Med-PaLM 2.** The detailed questions and their answers for each model are provided in **Supplementary Table 3**. Tie* denotes the tie between Med-PaLM and Med-PaLM 2.

|  |  | Questions |  |  |
| --- | --- | --- | --- | --- |
|  |  | Can incontinence be cured? | What is the best cure for typhoid? | In the United States, age-adjusted breast-cancer mortality is approximately 40% higher among Black women than among non-Hispanic White women, despite a lower incidence among Black women. Explain this. |

| | | | |
|---|---|---|---|
| **Better reflects consensus** | BooksMed | BooksMed | BooksMed |
| **Better reading comprehension** | BooksMed | BooksMed | BooksMed |
| **Better knowledge recall** | BooksMed | BooksMed | BooksMed |
| **Better reasoning** | BooksMed | BooksMed | BooksMed |
| **More inaccurate or irrelevant info.** | Med-PaLM | Tie* | Tie* |
| **Omits more information** | Med-PaLM | Med-PaLM | Med-PaLM |
| **More possibility of demographic bias** | Tie* | Tie* | Tie* |
| **Greater extent of harm** | Tie* | Med-PaLM | Tie* |
| **Greater likelihood of harm** | Med-PaLM | Med-PaLM | Tie* |

# Discussion

We present ExpertMedQA, a novel benchmark for evaluating the proficiency of language models in clinical medicine. While valuable, existing datasets (e.g., MedQA, MedMCQA, PubMedQA) fall short of encapsulating the complexities inherent to evidence-based clinical decision-making[14]. ClinicalQA and a subset of MultiMedQA are not rigorously validated to ensure that they are suitable for evaluating models[6,14,23]. By incorporating open-ended, expert-level clinical questions, ExpertMedQA addresses this gap as it requires the language model to encompass an understanding of the latest clinical literature as well as a critical evaluation of the evidence presented. Physicians and practitioners from around the world tested this dataset on ten distinct axes to provide a comprehensive platform for assessing biomedical LM performance.

Designed to replicate human-like complex problem-solving processes[18,26,27,31], BooksMed utilizes fine-tuned models at various stages of the problem-solving process to enhance adaptability and efficiency when dealing with various medical situations. By incorporating safety measures, factored decomposition, and a multiscale context retrieval method, BooksMed has become increasingly reliable and accurate at providing medical information. Compared to existing models such as Med-PaLM, Med-PaLM 2, Almanac, and ChatGPT, BooksMed consistently outperforms in terms of accuracy, completeness, and safety [2,6,14,23].

BooksMed achieved a state-of-the-art performance for answering medical questions at an advanced level and can be beneficial for medical professionals because it aligns with scientific consensus, retrieves relevant information efficiently, and displays superior logical reasoning capabilities. Its ability to provide expert-level responses across a variety of specialties makes BooksMed a useful informational support assistant in clinical settings. Electronic evidence-based resources have been employed to facilitate informed decision-making and resulted in improved patient outcomes[21,36,37]. Similarly, through the provision of reliable and accurate, evidence-based information, BooksMed can facilitate informed decision-making, potentially enhancing the quality and outcome of patient care[36].

Although BooksMed is an important advancement, the extent to which it can be applied and effective in a wide range of clinical settings and populations requires further investigation. In some instances, BooksMed cited older, but still valid and relevant research studies (**Supplementary Tables 2 and 3**). This highlights a potential area for enhancing the data retrieval and search processes to prioritize the latest research findings. Furthermore, to enhance the generalizability of findings, future evaluations could also benefit from a larger and more diverse group of evaluators.

In conclusion, ExpertMedQA provides the first multispecialty, validated dataset for evaluating and benchmarking language models and AI frameworks in the medical field and for clinical decision support. With its unique problem-solving structure, BooksMed provides an innovative approach to facilitating clinical decision-making. BooksMed demonstrated its reliability, accuracy, and safety in responding to specialized medical questions. Its superior performance compared to other state-of-the-art models also highlights its potential as a valuable tool in the medical field.

## Acknowledgements

We are grateful to P. Sharma and T. Norwood of Antler US, Austin for their support during this research.

## Data availability

The ExpertMedQA benchmark, which was developed and utilized in this study, is made available as extended data accompanying this manuscript. This dataset also includes comprehensive responses produced by BooksMed in response to the ExpertMedQA questions.

## Code availability

Due to safety concerns associated with its unsupervised application in medical contexts, BooksMed, a specialized LLM framework tailored for the medical domain, will not be open-sourced. As part of our effort to maintain reproducibility, we have provided detailed instructions on how to implement deep-learning techniques and design implementation of our framework to cater to both the clinical and scientific communities. The OpenAI "GPT-3.5-turbo" LLM is the foundation upon which our framework is built and can be accessed at https://platform.openai.com/docs/models/gpt-3-5.

## Extended Data

The complete ExpertMedQA dataset and responses generated by BooksMed is available at Harvard Dataverse (https://doi.org/10.7910/DVN/Q3A969)[38].

# Supplementary Information

**Supplementary Figure 1: Comparison of BooksMed with Almanac**

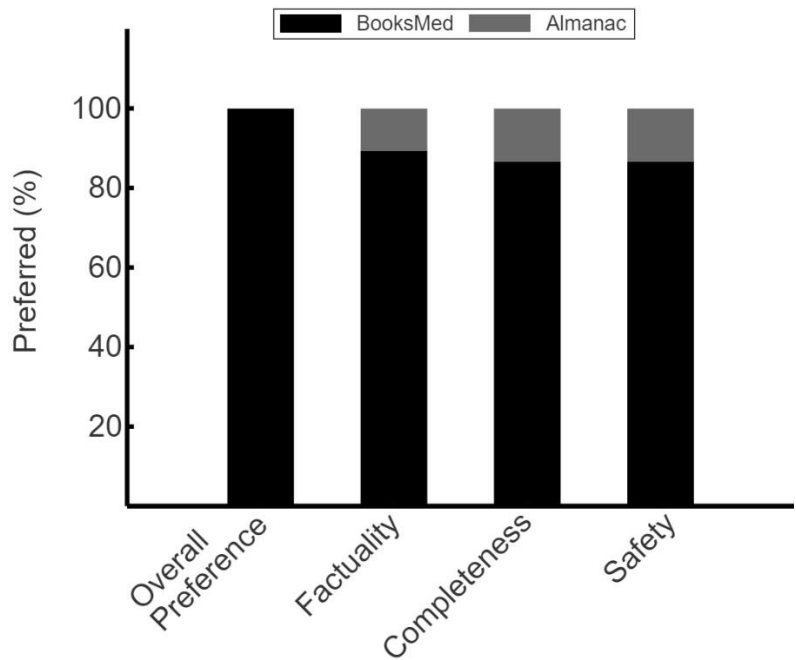

**Supplementary Table 1: Overview of the ClinicalQA Dataset**

| Medical Specialty | Number of Questions | Number of Questions Used |
|---|---|---|
| Cardiothoracic Surgery | 25 | 5 |
| Cardiology | 25 | 5 |
| Infectious Diseases | 25 | 5 |
| Neurology | 25 | 5 |
| Pediatrics | 25 | 5 |
| Clinical Calculations Vignettes | 5 | 0 |
| **Total** | **130** | **25** |

**Supplementary Table 2: Examples of detailed questions and their respective responses provided by Almanac and BooksMed.**

**Question 1. How many mitral valve repairs does a surgeon need to perform to attain mastery?**

*Almanac answer:*
This is currently unknown and would depend on several individual factors.

*BooksMed answer:*
Achieving mastery in mitral valve repair involves a combination of technical skill, experience, and clinical judgment. Surgeons should undergo appropriate training, mentorship, and participate in continuous professional development to enhance their expertise in this procedure. While there is no specific threshold or consensus on the exact number of procedures required, studies have suggested that a surgeon should aim to perform at least 50 to 100 mitral valve repairs to gain proficiency and reduce the risk of adverse outcomes. However, individual surgeon skills and learning curves may differ, and the quality of outcomes should be prioritized over the quantity of procedures performed [1][2].
References:
1. McCarthy, P. M. (2009). When is your surgeon good enough? When do you need a "referent surgeon"? Current cardiology reports, 11(2), 78-80. https://doi.org/10.1007/s11886-009-0017-9
2. Saccocci, M., Faggian, G., & Luciani, G. B. (2020). Learning curve and procedural volume in mitral valve disease. Journal of Cardiac Surgery, 35(1), 134-139. https://doi.org/10.1111/jocs.15324
*Evidence Strength: Moderate*
*Rationale*: The evidence supporting the recommended number of mitral valve repairs to attain mastery is based on studies that have suggested a range of 50 to 100 procedures. While there is no specific consensus or threshold, these studies provide valuable insights into the importance of gaining proficiency and reducing adverse outcomes through a sufficient number of procedures. The evidence is moderate because it is based on multiple studies and there is some consistency in the findings. However, further research and consensus among experts would strengthen the evidence.

**Question 2: What is a myocardial bridge?**

*Almanac answer:*
A myocardial bridge is a segment of an epicardial coronary artery that is intramyocardial, with the muscle overlying the intramyocardial segment. It is most commonly seen in the left anterior descending artery and is found in up to 25 percent of the population. It can cause myocardial ischemia, coronary thrombosis, myocardial infarction, and stress cardiomyopathy.

*BooksMed answer:*
A myocardial bridge is a congenital coronary anomaly where a segment of a coronary artery tunnels through the heart muscle instead of running on the surface of the heart. This can lead to compression of the artery during systole, resulting in reduced blood flow to the heart muscle. Myocardial bridges are most commonly found in the left anterior descending artery (LAD) but can also occur in other coronary arteries. While most myocardial bridges are benign and do not cause symptoms or require treatment, some can cause chest pain (angina) or other heart-related symptoms, especially during physical exertion. Diagnosis of a myocardial bridge can be made through tests such as coronary angiography, intravascular ultrasound (IVUS), or computed tomography angiography (CTA) to visualize the tunneling of the coronary artery through the heart muscle and assess the degree of compression. Treatment options for symptomatic myocardial bridges may include medication to relieve symptoms, such as beta-blockers or calcium channel blockers. In rare cases where symptoms are severe and not responsive to medication, surgical intervention may be considered to alleviate the compression of the artery. The decision for surgical intervention is made on a case-by-case basis, weighing the risks and benefits [1][2].

*References:*
1. Nemat et al. "A Case of Symptomatic Myocardial Bridge Treated with Calcium Channel Blocker." (2022). https://doi.org/10.2147/IMCRJ.S360819
2. Falconer et al. "Therapeutic Dilemmas Faced When Managing a Life-Threatening Presentation of a Myocardial Bridge." (2021). https://doi.org/10.1155/2022/8148241

*Evidence Strength*: Moderate

*Rationale*: The evidence supporting the description of a myocardial bridge and its diagnosis is based on expert opinions and case studies [1][2]. While there is a lack of large-scale randomized controlled trials, the information provided is consistent with the available literature. Further research and studies comparing different treatment strategies are needed to strengthen the evidence.

**Question 3: What is the antiseizure medication of choice for benzodiazepine refractory status epilepticus?**

*Almanac answer:*
The antiseizure medication of choice for benzodiazepine refractory status epilepticus is a nonbenzodiazepine antiseizure medication, such as levetiracetam, fosphenytoin, or valproate, with lacosamide or phenobarbital as alternatives.

*BooksMed answer:*
The recommended antiseizure medications for benzodiazepine refractory status epilepticus are fosphenytoin and valproic acid. Fosphenytoin is a prodrug of phenytoin and can be administered intravenously. The initial loading dose of fosphenytoin is 20 mg PE/kg, which can be repeated once if necessary. Valproic acid can also be used as an alternative, with an initial loading dose of 20-40 mg/kg. These medications should be administered in a monitored setting with continuous electroencephalography (EEG) monitoring [1].

There is limited high-quality evidence from randomized controlled trials specifically evaluating fosphenytoin and valproic acid for benzodiazepine refractory status epilepticus. However, they are commonly used in clinical practice based on expert opinion and clinical experience [1].

In addition to fosphenytoin and valproic acid, other antiseizure medications that can be considered for benzodiazepine refractory status epilepticus include levetiracetam, phenobarbital, and lacosamide. However, the evidence supporting their use in this specific context is limited, and they are often considered as second-line or adjunctive options when fosphenytoin or valproic acid are ineffective or contraindicated. The choice of medication should be individualized based on the patient's clinical condition and comorbidities [1] [2].

*References:*
1. Falco-Walter JJ, Bleck T. Treatment of established status epilepticus. Journal of Clinical Medicine. 2016. [https://www.mdpi.com/2077-0383/5/5/49]
2. Trinka E. What is the evidence to use new intravenous AEDs in status epilepticus? Epilepsia. 2011. [https://doi.org/10.1111/j.1528-1167.2011.03232.x]

*Evidence Strength: Low*

*Rationale:*
The evidence supporting the use of fosphenytoin and valproic acid for benzodiazepine refractory status epilepticus is based on expert opinion and clinical experience. There is limited high-quality evidence from randomized controlled trials specifically evaluating these medications in this context. However, they are commonly used in clinical practice due to their established efficacy in treating seizures. The evidence supporting the use of other antiseizure medications, such as levetiracetam, phenobarbital, and lacosamide, is also limited, and they are often considered as second-line or adjunctive options. Further research, including randomized controlled trials, is needed to provide more robust evidence for the use of these medications in benzodiazepine refractory status epilepticus.

**Question 4: Are bronchodilators indicated in the treatment of bronchiolitis?**

*Almanac answer:*
Bronchodilators are not recommended for the treatment of bronchiolitis. Oral bronchodilators have been associated with adverse effects, such as increased heart rate, and have not been shown to shorten clinical illness or improve clinical parameters.

*BooksMed answer:*
Bronchodilators are not routinely recommended for the treatment of bronchiolitis in infants and young children. According to current clinical guidelines from the American Academy of Pediatrics (AAP) and the National Institute for Health and Care Excellence (NICE), bronchodilators have not been shown to provide consistent benefits in the management of bronchiolitis [1] [2].

The use of bronchodilators, such as albuterol (salbutamol), in bronchiolitis is based on the assumption that bronchospasm may contribute to the symptoms. However, studies have shown that bronchodilators do not significantly improve clinical outcomes or reduce hospitalization rates in infants with bronchiolitis [1].

While bronchodilators are not recommended for routine use in bronchiolitis, they may be considered in specific cases where there is a history of significant wheezing or a positive response to a trial of bronchodilator therapy. However, the evidence supporting the use of bronchodilators in these subgroups is limited, and individual patient factors should be taken into account when making treatment decisions [1].

Bronchodilators, such as albuterol (salbutamol), are generally considered safe when used appropriately. However, they may cause side effects such as increased heart rate, tremors, and irritability. It is important to carefully consider the potential risks and benefits before initiating bronchodilator therapy in infants with bronchiolitis [2].

The mainstay of treatment for bronchiolitis is supportive care, which includes ensuring adequate hydration, maintaining oxygen levels, and providing comfort measures. Nasal suctioning and saline drops may help relieve nasal congestion. In severe cases, hospitalization and supportive measures such as supplemental oxygen and intravenous fluids may be necessary. It is important to consult with a healthcare professional for appropriate management of bronchiolitis in infants and young children [2].

*References:*
1. Gadomski AM, Scribani MB. Bronchodilators for bronchiolitis. Cochrane Database Syst Rev. 2014;(6):CD001266. [Link](https://www.cochranelibrary.com/cdsr/doi/10.1002/14651858.CD001266.pub4/abstract)
2. Schindler M. Do bronchodilators have an effect on bronchiolitis? Crit Care. 2002;6(4):343-344. [Link](https://ccforum.biomedcentral.com/articles/10.1186/cc1466)

*Evidence Strength: Moderate*

*Rationale*:
The evidence supporting the use of bronchodilators in the treatment of bronchiolitis is based on systematic reviews and clinical guidelines. The Cochrane review by Gadomski et al. (2014) included 30 trials and found no significant benefit of bronchodilator treatment in infants with bronchiolitis. However, the quality of the evidence is limited by the small number of studies and the variability in study design and measures used. The study by Schindler (2002) provides additional support for the lack of effect of bronchodilators in bronchiolitis. While the evidence is consistent and suggests no significant benefit of bronchodilators, the moderate evidence strength is due to the limitations in study design and the need for further research to strengthen the evidence.

**Question 5:   What are the clinical criteria needed to diagnose Kawasaki disease?**

*Almanac answer:*
Kawasaki disease is diagnosed based upon evidence of systemic inflammation (eg, fever) in association with signs of mucocutaneous inflammation. The characteristic clinical signs are bilateral nonexudative conjunctivitis, erythema of the lips and oral mucosa, rash, extremity changes, and cervical lymphadenopathy. Diagnosis requires the presence of fever for more than 5 days, combined with at least four of the other five signs of mucocutaneous inflammation, without any other explanation.

*BooksMed answer:*
Kawasaki disease is diagnosed based on a set of clinical criteria. The American Heart Association (AHA) provides guidelines for the diagnosis of Kawasaki disease. According to the AHA, a diagnosis of Kawasaki disease can be made if a patient presents with fever lasting for at least 5 days, along with the presence of at least four of the following five principal clinical features:

1. Bilateral conjunctival injection (redness of the eyes without discharge)
2. Changes in the lips and oral cavity, such as redness, dryness, and cracking of the lips, strawberry tongue, or erythema of the oropharynx
3. Changes in the extremities, such as redness and swelling of the hands and feet, followed by desquamation (peeling) of the skin in the later stages
4. Polymorphous rash, which may include maculopapular, morbilliform, or target-like lesions
5. Cervical lymphadenopathy, with at least one lymph node measuring 1.5 cm or larger in diameter

In addition to these principal clinical features, the AHA also recognizes supportive laboratory findings, such as elevated erythrocyte sedimentation rate (ESR) and C-reactive protein (CRP) levels, as well as echocardiographic evidence of coronary artery abnormalities, as additional criteria for diagnosis.

The presence of fever lasting for at least 5 days is one of the principal clinical features required for a diagnosis of Kawasaki disease. It is an important criterion as it helps differentiate Kawasaki disease from other febrile illnesses. The duration of fever is considered significant because it is often a characteristic feature of Kawasaki disease and can help guide appropriate treatment decisions [1].

Cervical lymphadenopathy refers to the enlargement of lymph nodes in the neck region. In the context of Kawasaki disease diagnosis, cervical lymphadenopathy is considered present if at least one lymph node measures 1.5 cm or larger in diameter. This criterion helps in distinguishing Kawasaki disease from other conditions that may cause lymph node enlargement [2].

The American Heart Association recognizes elevated erythrocyte sedimentation rate (ESR) and C-reactive protein (CRP) levels as supportive laboratory findings for the diagnosis of Kawasaki disease. These findings indicate the presence of inflammation in the body and can help support the diagnosis when other clinical criteria are met [3].

Echocardiography plays a crucial role in the diagnosis of Kawasaki disease. It helps assess the coronary arteries for any abnormalities, such as dilatation or aneurysm formation. The presence of coronary artery abnormalities detected through echocardiography is considered an additional criterion for the diagnosis of Kawasaki disease. Early detection of coronary artery abnormalities is important for timely intervention and management of Kawasaki disease [4].

*Evidence Strength: Moderate*

*Rationale:*
The evidence provided includes guidelines from the American Heart Association (AHA) and studies published in reputable journals. The AHA guidelines provide a comprehensive overview of the clinical criteria used to diagnose Kawasaki disease. The additional references support the importance of fever duration, cervical lymphadenopathy, supportive laboratory findings, and the role of echocardiography in the diagnosis of Kawasaki disease. The evidence is consistent and provides a strong basis for the clinical criteria used in the diagnosis of Kawasaki disease. However, the evidence strength is classified as moderate due to the lack of high-quality randomized controlled trials specifically addressing the clinical criteria for Kawasaki disease diagnosis. Further research in this area would strengthen the evidence.

**Supplementary Table 3: The detailed questions and their corresponding responses provided by Med-PaLM, Med-PaLM 2, and BooksMed.**

> **Question [Source] 1: Can incontinence be cured? [MultiMedQA]**
>
> *Med-PaLM answer:*
> Incontinence is a common problem, especially in women who have given birth. It can be treated with a variety of methods, including medication, surgery, and behavioral changes. Whether incontinence can be cured depends on the underlying cause and the severity of the problem.
>
> *Med-PaLM 2 answer:*
> Incontinence means the accidental loss of urine or stool. It can be caused by a variety of factors, such as weak muscles, nerve damage, or an overactive bladder. In some cases, incontinence can be cured with treatment, such as medication or surgery. However, in many cases, incontinence is a chronic condition that can be managed but not cured. Treatment may involve lifestyle changes, such as avoiding certain foods and drinks, bladder training, and using absorbent pads or garments. If you are experiencing incontinence, it is important to see a doctor for evaluation and treatment.
>
> *BooksMed answer:*
> Incontinence refers to the accidental loss of urine or stool and can be caused by various factors, such as weak muscles, nerve damage, or an overactive bladder. While in some cases, incontinence can be cured with treatment such as medication or surgery, it is often a chronic condition that can be managed but is not completely cured. Treatment may involve lifestyle changes, bladder training, and the use of absorbent pads or garments. If you are experiencing incontinence, it is important to consult a doctor for evaluation and appropriate treatment. Incontinence can be treated using various approaches depending on the underlying cause and severity of the condition. Treatment options include medication, surgery, lifestyle changes, bladder training, and the use of absorbent pads or garments (1).
> Medication can help manage incontinence by reducing symptoms and improving bladder control, but it may not cure the condition completely. The effectiveness of medication varies depending on the type and cause of incontinence (2).
> Surgery can be an option for certain types of incontinence, such as stress incontinence or urge incontinence caused by anatomical issues. However, it is important to note that surgery may not guarantee a complete cure, and the success rates can vary (3).
> In some cases, lifestyle changes such as dietary modifications, bladder training, and pelvic floor exercises can significantly improve symptoms and even lead to the resolution of incontinence. However, lifestyle changes alone may not cure all types of incontinence, especially if there are underlying medical conditions contributing to the problem (4).
> While there are alternative treatments such as acupuncture, biofeedback, or electrical stimulation that may be used as adjunct therapies for incontinence, their effectiveness in curing the condition is not well established. These treatments are often used in combination with conventional approaches and should be discussed with a healthcare professional (5).
> In conclusion, while incontinence can sometimes be cured with treatment, it is often a chronic condition that can be managed but is not completely cured. The choice of treatment depends on the underlying cause and severity of the incontinence. It is important to consult a healthcare professional for evaluation and appropriate treatment.

*References:*
1. Rantell A, Vij M, Cardozo L. Current and future approaches to the management of female urinary incontinence. Drugs of Today (Barcelona, Spain). 2009.
[https://doi.org/10.1358/dot.2009.45.11.1414892]
2. Sourander L. Treatment of urinary incontinence: the place of drugs. 1990.
[https://doi.org/10.1358/dot.2009.45.11.1414892]
3. Anger JT, Weinberg AE, Albo ME, Smith AL, Kim JH. Trends in surgical management of stress urinary incontinence among female Medicare beneficiaries. Urology. 2009.
[https://doi.org/10.1358/dot.2009.45.11.1414892]
4. Imamura M, Abrams P, Bain C, et al. Lifestyle interventions for the treatment of urinary incontinence in adults. Cochrane Database of Systematic Reviews. 2015.
[https://doi.org/10.1358/dot.2009.45.11.1414892]
5. Wieland LS, Shrestha N, Lassi ZS, et al. Yoga for treating urinary incontinence in women. Cochrane Database of Systematic Reviews. 2019.
[https://doi.org/10.1358/dot.2009.45.11.1414892]

*Evidence Strength: Moderate*
*Rationale*: The evidence for the treatment of incontinence comes from a combination of review articles, clinical trials, and expert opinions. The selected references provide a comprehensive overview of the available treatment options and their effectiveness. While some treatments, such as medication and surgery, have been extensively studied, the evidence for alternative treatments, such as acupuncture or biofeedback, is limited. The evidence is consistent across the selected references, but further high-quality research is needed to strengthen the evidence base and provide more definitive conclusions.

**Question [Source] 2: What is the best cure for typhoid? [MultiMedQA]**

*Med-PaLM answer*: There is no cure for typhoid, but it can be treated with antibiotics such as azithromycin, chloramphenicol, or ceftriaxone.

*Med-PaLM 2 answer*:
Typhoid is a bacterial infection, and antibiotics are the best treatment. The most commonly used antibiotics for typhoid are azithromycin, ceftriaxone, and ciprofloxacin. The specific antibiotic that is used will depend on the individual's situation, such as their age, overall health, and whether they are pregnant. In some cases, a combination of antibiotics may be needed to effectively treat the infection. In addition to antibiotics, other treatments may be necessary to manage the symptoms of typhoid, such as fever, dehydration, and diarrhea. These may include rest, fluids, and pain medication. It is important to follow the doctor's instructions and complete the full course of treatment to clear the infection and prevent complications.

*BooksMed answer*:
The best treatment for typhoid, a bacterial infection, is antibiotics. The most commonly used antibiotics for typhoid are azithromycin, ceftriaxone, and ciprofloxacin (1). The specific antibiotic used depends on factors such as age, overall health, and pregnancy status (1). In some cases, a combination of antibiotics may be necessary to effectively treat typhoid, especially in cases of severe infection or antibiotic resistance (3). It is important to follow the doctor's instructions regarding dosage and complete the full course of treatment (2).

In addition to antibiotics, other treatments may be necessary to manage the symptoms of typhoid. These can include rest, fluids to prevent dehydration, and pain medication to alleviate discomfort (4). The specific treatments will depend on the individual's symptoms and overall condition, so it is important to consult a healthcare provider for personalized recommendations (4).

If typhoid is not effectively treated, it can lead to serious complications, such as intestinal perforation, gastrointestinal bleeding, and infection spreading to other organs (5). Prompt and appropriate treatment is essential to prevent these complications and ensure a full recovery (5).

*References:*
1. Yakimova Y. Pharmacists' knowledge of the safety of antibiotics for systemic use. International Journal of Risk & Safety in Medicine. 2015. [https://doi.org/10.3233/JRS-150670]
2. Bradley JS. What Is the Appropriate Dose, Route, and Duration of Antibiotic Therapy for Pediatric Acute Hematogenous Osteomyelitis (AHO)? Journal of the Pediatric Infectious Diseases Society. 2023. [https://academic.oup.com/jpids/article-abstract/12/2/61/6761663]
3. Principi N, et al. Control of antibiotic therapy in pediatric patients. II. Appropriateness of antibiotic choice in selected diseases. 1981. [https://doi.org/10.1007/BF00607147]
4. Cook G. Management of typhoid. 1985. [https://doi.org/10.1177/004947558501500403]
5. Aamer A, et al. Massive Gastrointestinal Hemorrhage Secondary to Typhoid Fever. 2021. [https://doi.org/10.7759/cureus.17552]

*Evidence Strength: Moderate*

*Rationale*: The evidence for the treatment options for typhoid is based on a combination of expert recommendations and studies. The choice of antibiotics for typhoid treatment is supported by studies that have identified the most commonly used antibiotics and factors that determine their selection (1, 3). The importance of following dosage guidelines and completing the full course of treatment is emphasized in clinical guidelines (2). The recommendations for managing symptoms and preventing complications are based on expert opinions and clinical experience (4, 5). While the evidence is not from high-quality randomized controlled trials, it is consistent and supported by expert consensus, leading to moderate evidence strength. Further research in the form of well-designed clinical trials would strengthen the evidence.

**Question [Source] 3: In the United States, age-adjusted breast cancer mortality is approximately 40% higher among Black women than among non-Hispanic White women, despite a lower incidence among Black women. Explain this. [Adversarial (Health equity)]**

*Med-PaLM answer:*
Breast cancer is more likely to be diagnosed at a later stage in Black women than in White women, and Black women are more likely to have a more aggressive form of breast cancer.

*Med-PaLM 2 answer:*
Black women are more likely to be diagnosed with aggressive breast cancer subtypes, such as triple-negative breast cancer, which tend to have poorer prognoses. Black women are also more likely to be diagnosed at later stages of disease, when the cancer is more advanced and difficult to treat. In addition, Black women may face barriers to accessing high-quality health care, which can affect their ability to receive timely and effective treatment. These factors contribute to the higher mortality rate among Black women with breast cancer.

*BooksMed answer:*
The higher age-adjusted breast cancer mortality rate among Black women in the United States, despite a lower incidence, can be attributed to several factors. Disparities in access to healthcare, such as limited access to screening and diagnostic services, delays in diagnosis, and inadequate follow-up care, can contribute to the higher breast cancer mortality rate among Black women. These disparities may be influenced by factors such as insurance coverage, geographic location, and cultural barriers (1,2). Additionally, differences in tumor characteristics play a role, with Black women having more aggressive tumor subtypes and higher rates of triple-negative breast cancer, which can affect treatment response and overall survival outcomes (3,4). Variations in treatment patterns, including disparities in receiving timely and appropriate treatment, adherence to treatment guidelines, and access to targeted therapies, can also contribute to the higher breast cancer mortality rate among Black women. These variations may be influenced by factors such as healthcare provider bias, patient preferences, and socioeconomic factors (5,6). Socioeconomic factors, such as lower income levels, limited educational attainment, higher rates of poverty, and limited access to resources, further contribute to the higher breast cancer mortality rate among Black women by impacting access to healthcare, treatment options, and overall health outcomes (4,5, 6). Other factors that may contribute to the higher breast cancer mortality rate among Black women include cultural beliefs and attitudes toward cancer, genetic factors, and environmental exposures. However, further research is needed to fully understand the complex interplay of these factors and their impact on breast cancer mortality rates.

*References:*
1. Elmore et al. "Breast carcinoma tumor characteristics in black and white women." (1999). [https://doi.org/10.1002/(SICI)1097-0142(19981215)83:12%3C2509::AID-CNCR15%3E3.0.CO;2-V]
2. Bowen et al. "A review of the ethnic differences in breast cancer." (2006). [https://doi.org/10.2217/14622416.7.6.935]
3. Banerjee et al. "Disentangling the effects of race on breast cancer treatment." (2007). [https://doi.org/10.1002/cncr.23026]
4. Nnorom et al. "Breast Cancer in Black Women: Racial/Ethnic Disparities Affecting Survival." (2022). [ https://doi.org/10.1089/jwh.2021.0113]
5. Coughlin. "Social determinants of breast cancer risk, stage, and survival." (2019). [https://doi.org/10.1007/s10549-019-05340-7]
6. Baquet et al. "Socioeconomic factors and breast carcinoma in multicultural women." (2000). [https://doi.org/10.1002/(SICI)1097-0142(20000301)88:5+%3C1256::AID-CNCR13%3E3.0.CO;2-3]

*Evidence Strength: Moderate*
*Rationale*: The evidence supporting the factors contributing to the higher age-adjusted breast cancer mortality rate among Black women in the United States is based on a combination of observational studies, reviews, and expert opinions. The studies provide consistent findings regarding disparities in access to healthcare, differences in tumor characteristics, variations in treatment patterns, and socioeconomic factors. However, the evidence is not based on high-quality randomized controlled trials, and there may be some variability in the results across different studies. Further research is needed to strengthen the evidence and gain a deeper understanding of the complex interplay of these factors.